\documentstyle [12pt]{article}
\newcommand {\ignore} [1] {}
\newtheorem{theorem}{Theorem}[section]

\newtheorem{definition}{Definition}[section]

\newtheorem{claim}[theorem]{Claim}

\begin{document}

\title{Rational Competitive Analysis}

\author{Moshe Tennenholtz
\thanks{Permanent address: Faculty of Industrial Engineering and Management, 
Technion--Israel Institute of Technology, Haifa 32000, Israel}\\
 Computer Science Department\\
 Stanford University \\
  Stanford, CA 94305}
\date{}
\maketitle

\begin{abstract}
Much work in computer science has adopted competitive analysis as a 
tool for decision making under uncertainty. In this work we extend 
 competitive analysis to the context of multi-agent systems. Unlike
classical competitive analysis where the behavior of an agent's environment
is taken to be arbitrary, we consider the case where an agent's environment
consists of other agents. These agents will usually obey some (minimal)
rationality constraints. This leads to the definition of {\em rational
competitive analysis}. We introduce the concept of rational competitive
analysis, and initiate the study of competitive analysis for
multi-agent systems. We also discuss the application of rational 
competitive analysis 
 to the context of bidding games, as well as to the classical one-way
trading problem.
\end{abstract}

\section{Introduction}

Competitive analysis is a  central tool for the design and analysis 
of algorithms and protocols for decision making under uncertainty 
\cite{BorElYa}. 
It is a well studied and widely applicable approach that fits the framework 
of qualitative decision-making in AI (see e.g. 
\cite{DT97,BraTenjour}). 
The competitive analysis approach attempts to minimize the ratio
between the payoff an  agent obtains and the payoff 
 he could have obtained had
he known the behavior of the environment. For example, consider the
following trading problem (see \cite{BorElYa}, Chapter 14). 
An agent who holds \$100 may wish to exchange them for British pounds. At each point in time, e.g. every minute in between 8AM and 4PM,
an exchange ratio
 of dollars
and pounds is announced. This ratio changes dynamically, and in
an unpredicted
manner.
The agent would need to choose the time in which it will trade his
\$100 for pounds.
Notice that if
the agent would have known the sequence of exchange rates, $e$,
then  he could have chosen a strategy $o(e)$ that maximizes his
payoff. If the agent chooses a  strategy $s$, then we
can compute the ratio of the payoffs obtained by $o(e)$ and $s$.
We can do similarly for every sequence of exchange rates $e'$. 
Based on this, we can compute 
the highest (i.e. worst) ratio, over all possible sequences, that might 
be obtained when we compare optimal strategies to $s$. This ratio is
denoted by $R(s)$.  According to the competitive analysis approach,
the agent will apply the {\em competitive ratio} decision criterion:
 he will choose
  a strategy $s$ for which $R(s)$ is
minimal. This decision criterion may be quite helpful when we
lack probabilistic  assumptions about the environment. For
example, assume that the minimal value of $R(\cdot)$, which
 is obtained by  some
strategy $s$, is 2. Then, by selecting $s$,
 the agent guarantees himself a payoff which is at
least half of the optimal payoff that he could have obtained had
he known the actual environment behavior.

The competitive ratio has also an additive variant (also termed 
minimax regret \cite{Milnor}), where we replace the term "ratio"
by the term "difference" in the definition of $R$. So, in our
example, if a strategy $s$ that minimizes $R(\cdot)$ obtains a 
competitive difference/regret of 20, 
then this implies that by performing $s$ the agent
gets a payoff (e.g. worth in British pounds) which is at most 20
lower than what he  could have obtained had he known the behavior of the
environment. In the sequel,  we will use the additive version of
the competitive ratio decision criterion.

 Competitive analysis 
 has been applied to a variety of classical problems 
 in computer science, 
 such as the k-server problem \cite{KoutsoupiasPapadimitriou} and 
paging \cite{FiatKarpLuby}, as well as to 
more  general algorithmic 
problems \cite{BorodinLinialSaks}.
 In all of these studies the environment that the agent acts in is
 non-strategic,  and therefore does not assume to follow any "rational"
 behavior. In this paper we extend the concept of competitive
 analysis to the context of multi-agent systems. In a multi-agent system
 the
 environment in which an agent takes his decision consists of
 other "rational" agents. Following 
previous work on competitive analysis, our
 approach is non-Bayesian and normative;
 we would like to find a decision rule for
 the agent that will rely  as little as possible on assumptions about
 the behavior of his environment. Therefore, we adopt the requirement 
 that agent $A$ should
 rule out a behavior $b_1$ of agent $B$ only if $b_1$ is dominated
 by another behavior, $b_2$, of that agent.
 Dominated behaviors are purely irrational in any decision
 making model. The agent will choose his behavior according to the
 competitive ratio decision rule. However, he should consider only
 rational behaviors of the other agents; a behavior of an 
agent will be considered
 irrational if and only if it is dominated by another behavior of it.

 In section 2 we describe bidding games, a family of games that
 will serve us for the illustration of the basic concepts developed
 in this paper. Bidding games are representatives of k-price
 auctions, a central class of economic mechanisms \cite{kpricegeb}.
 In section 3 we present a competitive analysis of bidding games.
 In section 4 we introduce rational competitive analysis, a new
 tool for normative decision making, that generalizes competitive
 analysis to the context of rational environments, and apply it  
to bidding games. 
In section 5 we consider repeated (multi-stage) games. We present several
 results on the relationships 
 between (rational) competitive analysis
of repeated games and the competitive analysis 
of the particular (one-shot) games they consist of.
Then, in section 6,  we discuss and study  variants and modifications
 of the one-way trading problem, using rational competitive 
analysis.  In particular, we 
 study the multi-agent one-way trading problem.

\section{Bidding games}

We start by recalling the general definition of a
(strategic form) game.

\begin{definition}
A {\em game} is a tuple $G=\langle
N=\{1,2,\ldots,n\},S=\{S_i\}_{i=1}^n, \{U_i\}_{i=1}^n \rangle$,
where $N$ is a set of $n \geq 2$ players, $S_i$ is the set of
strategies available to player $i$, and $U_i: \Pi_{j=1}^n S_j
\rightarrow R$ is the utility function of player $i$.
\end{definition}

In a game, each player selects a strategy from a set of available strategies.
The tuple of strategies selected, one by each player,  determines
 the payoff of each of the agents%
\footnote{Here and elsewhere we use the terms player and agent 
interchangeably.} (as prescribed by the utility functions). 

In a bidding game, a center attempts to obtain a service from a set of
potential suppliers. Each such supplier has a certain cost associated
with that service. This cost is taken to be an integer between $K-T$
and $K$, where $K$ and $T$ are w.l.o.g integers, $K>T > 0$.  Each agent
will offer his service and ask for  a payment in the range in between
$K-T$ and $K$. We will associate the request for payment of 
 $K-i$ with the
integer $i$, where $0 \leq i \leq T$.  The center will choose 
as the service provider the supplier 
 with the lowest asking price.  There are various ways
for determining the payment to that agent; in particular, the agent can be
paid his asking price, the second lowest asking price, 
or the third lowest
asking price. We assume  that the costs for providing the service by
each of the agents are common knowledge among them, although
the center might not be familiar with these costs.
 Although this is quite natural
for the above procurement problem, other assumptions can be treated
similarly.

Our definition of bidding games will capture 
the above, by considering
a fully isomorphic situation, namely: the
auctioning of a good.
The good is held by the center.
Each agent has a valuation (i.e. maximal willingness to pay) for the good.
 Each agent needs to decide on his bid. The center will 
allocate the good to the agent with the highest bid (rather than
to the agent with the lowest asking price, 
as in the isomorphic procurement problem).

Formally, we have:

\begin{definition}
Given a set of $n$ players, and an integer $T >> 1$,
 a bidding game is determined by
 the tuple
B=$(x_1,\ldots,x_n,k)$ where $x_i={l_i \over T}$ for some integer
$0 \leq l_i \leq T$, and $1 \leq k \leq n$ is an integer.  
Player $i$'s strategy in $B$ is a decision about $b_i\in [0,T]$.
Given a strategy profile $b=(b_1,b_2,\ldots,b_n)$ denote by $b_{[i]}$
the $i$-th order statistic of this tuple.
Let $M(b)$ be the number of elements of $b$ that equals $b_{[1]}$.
Then, $U_i(b)= {1 \over M(b)} (x_i - {b_{[k]} \over T})$ if
$b_i=b_{[1]}$, and $U_i(b)=0$ otherwise.
\end{definition}

In the above formalism, $x_i$ is the valuation of agent $i$
(that is normalized to the interval [0,1]), while $b_i$ denotes
the bid made by agent $i$.
The
payment made by the winner 
is determined by the parameter $k$. If $k=1$ we get the
standard high-bid wins (or first-price) auction; if $k=2$ then we
get the famous Vickrey (second-price) auction, while if $k=3$ we
get the case of third-price auctions.%
\footnote{Third-price auctions have been shown to have
appealing properties in the context of Internet Auctions 
\cite{kpricegeb}.}

 For ease of presentation we will assume that $ 2 \leq l_i <  T$
for every $1 \leq i \leq n$, that $i \neq j$ implies $l_i \neq
l_j$, and that $T \geq n$.

\section{Competitive Analysis}

In a game, agent $i$  is facing an environment that consists of
the other agents. The actions to be selected by these agents are
not under the control of $i$. Following the literature on
competitive analysis, the competitive ratio decision rule
may be used in order
to choose an action for that agent.

\noindent{\bf Definition 3.1} 
%
{\em 
Given a game $G$, and a strategy profile $s=(s_1,s_2,\ldots,s_n)
\in \Pi_{j=1}^n S_j$, the regret of player $i$ is given by
$Reg_i(s_i,s_{-i})=max_{t \in S_i}
U_i(s_1,\ldots,s_{i-1},t,s_{i+1},\ldots,s_n) - U_i(s)$. A
strategy $s \in S_i$ is a competitive strategy for agent 
$i$ if $s \in argmin_{t
\in S_i} max_{q \in S_{-i}} Reg_i(t,q)$, where $S_{-i}$ denotes
the possible strategy profiles of players in $N \setminus \{i\}$.}

Given the above definition we are interested in applying
competitive analysis to bidding games. We now present three
claims about competitive analysis of  bidding games. 
These claims are associated with the competitive analysis of 
 1st,2nd, and 3rd-price auctions, respectively. 

\begin{claim} Given the bidding game
B=$(x_1,\ldots,x_n,1)$, a competitive strategy for agent $i$ yields
a regret value of $\alpha \over T$, where $\alpha$ 
equals the upper integer value of ${{l_i-1} \over {2}}$.

\end{claim}

\noindent{Basic idea behind proof:} Agent $i$ can lose by
submitting a bid that is higher than his valuation. On the other
hand, by submitting a bid that is below ${l_i}-1$ agent $i$ might
lose, since agent $j \neq i$ might submit $l_i-1$ as a  winning bid. 
 Since agents may submit the bid 0,
agent $i$ will minimize his regret by submitting a bid
that equals (the upper integer value of) half of the difference
between  $l_i-1$ and 0.

\begin{claim} Given~the~bidding game B=$(x_1,\ldots,x_n,2)$ a competitive strategy
for agent $i$ yields a zero
regret.

\end{claim}

\noindent{Basic idea behind proof:} Here the optimal strategy for
an agent, regardless of what the others do, is to send his actual
valuation as his bid; this is a well known property of the Vickrey
auction \cite{Wolfstetter}. As a result we get a regret of 0.

\begin{claim} Given the bidding game
B=$(x_1,\ldots,x_n,3)$,  and assume w.l.o.g that $x_1 > x_2 > \cdots > x_n$,
then  agent $j$'s competitive strategy is to send the bid
${min(2l_j,T)}$.

\end{claim}

\noindent{Basic idea behind proof:} Given that agents may submit
the bid 0, agent $j$ might reach a regret of ${{l_j} \over T}$ if he is not
the winner. Submitting however a bid that is higher than $2l_j$
may also lead to a regret of ${{l_j} \over T}$, given that the agents may
submit $2l_j$ as their bids. Combining these
observations, we
get that submitting the bid ${{min(2l_j,T)}}$ is the
competitive strategy.

\section{Rational Competitive Analysis}

Although competitive analysis is a most powerful concept from a
non-Bayesian 
normative perspective, it may be quite restrictive when we consider
decision-making in multi-agent systems. Following the spirit of competitive
analysis for normative decision making,
we refrain from using probabilistic  assumptions and
 game-theoretic equilibrium analysis.%
\footnote{The debate about whether competitive ratio and non-Bayesian
decision making are expressive or useful for normative or descriptive
objectives is beyond the scope of this paper; see \cite{BTJACM99} for sound and 
complete
axiomatization of the competitive ratio decision criterion.}
 However, one can
still improve on the use of competitive analysis by considering 
minimal rationality requirements.

\noindent{\bf Definition 4.1} 
{\em 
Given a game $G=\langle N=\{1,2,\ldots,n\},\{S_i\}_{i=1}^n,\{U_i\}_{i=1}^n
\rangle$, we say that a strategy $s_i \in S_i$ weakly dominates a strategy
$s_i^{'} \in S_i$ if $U_i(s_i,t) \geq
U_i(s_i^{'},t)$ for every strategy profile $t$ of the players in
$N \setminus \{i\}$, and there exists such strategy profile $t'$ for which
$U_i(s_i,t') >
U_i(s_i^{'},t')$. A strategy $s \in S_i$ will be called rational is there
is no other strategy $\bar{s} \in S_i$ that weakly dominates it.
Given a game $G$, the set of rational strategies for player $i$ will
be denoted by $Rat(S_i)$.}

In any reasonable model agents will choose only from the set of
non-dominated strategies. Our idea is therefore to combine the
powerful idea of competitive analysis and this minimal
requirement of rationality, in order to re-introduce competitive
analysis into the framework of multi-agent systems.

\noindent{\bf Definition 4.2} 
{\em 
A strategy $s \in S_i$ is a rationally competitive strategy if $s
\in argmin_{t \in S_i} max_{q \in Rat(S_{-i})} Reg_i(t,q)$, where
$Rat(S_{-i})$ denotes the possible rational strategy profiles of
players in $N \setminus \{i\}$, i.e. each player $j \in N
\setminus \{i \}$ chooses its strategy from $Rat(S_j)$.}


Basically, a rationally competitive strategy applies the
competitive ratio decision criterion, while taking into account only
rational activities of the environment. As the following
claims illustrate, rational competitive analysis introduces 
 an improved approach to 
normative decision making.

\begin{claim}
Given the bidding game
B=$(x_1,\ldots,x_n,1)$,  a rational competitive strategy for
agent $i$ yields a regret of $\alpha \over T$, where $\alpha$ 
 equals the upper integer 
value of ${min(l_i,max_{j; j\neq i} l_j)-2} \over {2}$.
\end{claim}

{\noindent Basic idea behind proof:} We observe that any strategy
that tells agent $j$ to submit a bid which is greater than or equals to
his valuation is dominated by the strategy of submitting his
valuation minus 1. Given our assumptions about the possible valuations,
all other strategies, excluding the strategy of
submitting the bid 0, are not dominated. As a result, from the
perspective of agent $i$, if his valuation is the highest one,  he
will minimize his regret if he will make  a bid that is half of the
distance between
$max_{j; j\neq i} l_j - 1$ and 1. If agent $i$'s valuation is not the
highest one then he will minimize his regret (again, taking into
account the assumptions on possible valuations) if he will make a bid
in between $l_i-1$ and 1.

Notice that rational competitive analysis allows us to improve upon the 
type of reasoning carried out in claim 3.1. Technically, 
in the case of a bidding game with $k=1$, rationality 
implies that we need to take the minimum between $l_i$ and the highest other 
agents' valuation in our analysis, rather than consider  $l_i$ only.

\begin{claim}
Given the bidding game B=$(x_1,\ldots,x_n,2)$
 a rational competitive strategy for agent $i$ yields a zero
regret.

\end{claim}

As we can see, unlike the major effect of the rationality
assumption in the case of a first-price auction, there is no change
in the analysis in the case of a second-price auction. In the case of a
third-price auction, we see again the effect of the revised
notion:

\begin{claim}
Given the bidding game B=$(x_1,\ldots,x_n,3)$,  and assume
w.l.o.g that $x_1 > x_2 > \cdots > x_n$, then a rational
competitive strategy for agent $j \ (j=1,2)$, is  to submit the
bid $min(2l_j-l_{[3]},T)$, where $l_{[3]}$ corresponds to the
3rd highest $x_k$; a zero-regret rational competitive  strategy
for agent $i, 3 \leq i \leq n$, is to submit $l_i$.

\end{claim}

\noindent{Basic idea behind proof:} First, observe that any
strategy where the agent submits a bid that is below that agent's
valuation is dominated by the strategy that tells him to submit
his actual valuation as his bid. As a result, for agents $3,4,\ldots,n$
 there is a 0 regret in
submitting their  actual valuations as their bids.
 Let us assume that agent $i$ (where $i$ is 1 or 2)
submits a bid, then it can lose ${{l_i-l_{[3]}} \over T}$ if it turns out not
be the highest bidder (since agent $j$ submits a higher bid). 
 On the other hand, by submitting the bid $b_i > l_i$ a loss
of ${{b_i-l_i} \over T}$ may be caused, since (from the perspective of agent $i$) 
two other agents may submit the bid $b_i$. 
 This implies that the bid
$min(2l_i-l_{[3]},T)$ will minimize this agent's regret.

As we can see, in the case of $k=3$ as well, rational competitive analysis 
for bidding games leads to an improved normative approach to decision making. 
In particular, the competitive strategy of Claim 3.3 specifies a too high bid, 
and is not a rationally competitive strategy; as a result, it fails to 
serve in a multi-agent context.  

\section{Rational competitive analysis in repeated games: folk theorems}

We first recall the notion of finitely repeated games \cite{FudTir}.

\begin{definition}
Given an integer $l >0$ and a game $G=\langle
N=\{1,2,\ldots,n\},S=\{S_i\}_{i=1}^n, \{U_i\}_{i=1}^n \rangle$,
a {\em repeated game} $RG=(G,l)$ with respect to $G$ is a game where
$G$ is repeatedly played $l$ times. $RG$ consists of the
following strategies and utility functions:
a strategy of agent $i$ in $RG$ determines the strategy of $G$ to be taken
by $i$ in the $k$-th iteration of $G$, as a function of the history of
strategies of $G$
selected by the others in iterations $1,2,\ldots,k-1$.
Given a tuple of strategies of $RG$, one for each agent, the payoff
for agent $i$ is the sum of its payoffs along the $l$ iterations.
A sub-game of a repeated game $RG$ is a repeated game that starts from
iteration $ 1 \leq q \leq l$ of $RG$ and consists of $l-q+1$ iterations.
A (rationally) competitive strategy in $RG$ is a strategy
that is a (rationally) competitive strategy at each of the sub-games
of $RG$.%
\footnote{This definition is in the spirit of sub-game perfect equilibrium
in game theory.}
\end{definition}

Repeated games have been of much interest in the game-theory literature, 
due to the fact they enable to study agents' actions as a function
of past events and other agents' actions.
The study of repeated games is central to the
 understanding of
basic issues in coordination and cooperation (e.g. \cite{Axelrod}),  as well
as for the study of  learning in games (e.g. \cite{FudLevlearn}).

One of the central challenges for  the study of repeated games is to
establish general theorems (titled folk-theorems)
that explain/recommend  behavior in
these (repeated) games by means of solution concepts for
the games they consist of. In our case, it would be of interest
to understand what will be a rationally competitive strategy in a
repeated game, and try to relate it to the competitive analysis
of the simple one-shot game that takes place at each iteration.

We now present a general result about competitive analysis in
repeated games. For ease of presentation we will assume that $G$
is a two-player game, where all payoffs are distinct. We will also
assume w.l.o.g that  all payoffs are non-negative.  Given a
repeated game $(G,l)$, let us denote the highest payoff for agent
$i$ in $G$ by $h_i(G)$, and the second highest payoff of agent
$i$ in $G$ by $sh_i(G)$.

\begin{theorem}

Given a repeated game $(G,l)$ and assume that for each agent $i$
 $h_i(G) \geq 2 \cdot sh_i(G)$, then
a rationally competitive strategy for agent $i$
in the game $(G,l)$ is obtained
 by performing the competitive strategy of it in $G$ on iterations
 $1,2,\ldots,l-1$ and performing the rational competitive strategy of
it in $G$ on the
last iteration.
\end{theorem}

\noindent{Basic idea behind proof:} 
From the
perspective of agent $i$, assuming we are at stage $k < l$, the
selection of any strategy $s$ of $G$ by $j$ can be complemented to
a non-dominated strategy of $j$; this non-dominated strategy will tell
$j$ to choose the strategy associated with $h_j(G)$ in stages 
$k+1,\ldots,l$.  The reason
that the resulting strategy  is not dominated is that $j$
considers the strategy where $i$ will also choose in stages
$k+1,\ldots,l$ the  strategy (of his) in $G$ that corresponds to $h_j(G)$,
and does it only if in stage $k$ agent $j$ chooses $s$; in addition,
according to this strategy $i$ will choose the strategy that corresponds
to $sh_j(G)$ is stage $k$.
This
implies that agent $i$ should consider at stages $1,2,\ldots,l-1$
all possible strategies of agent $j$ in $G$. In the last stage
agent $i$ is no longer subject to the above considerations and
will choose the rationally competitive strategy of $G$.

The above theorem shows a strong connection between competitive analysis
in repeated games and competitive analysis in simple single-shot games.
As it turns out, this connection can be 
further generalized to a much richer context:

\begin{definition} Let $\bar{G}=(G_1,G_2,\ldots,G_m)$ 
be a sequence of games 
where $N$ is the set of players in each of the games in the sequence,
 and game $G_i$ is played in iteration $i$.
The strategy of agent $t$ in $\bar{G}$ determines its strategy in
$G_i$, $1\leq i \leq m$,
as a function of the strategies of  $G_j$, $1 \leq j < i$, selected
 by the other
agents in previous iterations.
Given a tuple of strategies of $\bar{G}$, one for each agent, the
payoff of agent $i$ is taken as the sum of its payoffs in the $m$
iterations.

\end{definition}

\begin{theorem}

Given a sequence of games $\bar{G}=(G_1,G_2,\ldots,G_m)$ where
$N$ is the set of players in each of the games in the sequence,
 and game $G_j$ is played in iteration $j$,
and assume that
 $h_i(G_k) \geq 2 \cdot sh_i(G_l)$ for every $1 \leq k,l \leq m$, 
and for every agent $i$, then
a rationally competitive strategy for agent $i$
in the game $\bar{G}$ is obtained
 by performing the competitive strategy of $G_j$ in iterations
 $1,2,\ldots,l-1$ and performing the rational competitive strategy of
$G_m$ in the
last iteration.
\end{theorem}

The above theorem can be generalized into a situation where $n$
games from among the set of games $\{G_1,G_2,\ldots,G_m\}$ are
executed in some random order (with possible repetitions).
Formally, this can be captured by the following definition and
theorem:

\noindent{\bf Definition 5.3} 
{\em 
Given a set of games $G=\{G_1,G_2,\ldots,G_m\}$, a random game
with respect to $G$,  $\bar{G}$,  is a sequence of $n$ games
$(g_1,g_2,\ldots,g_n)$, where $g_i \in G \ (1 \leq i \leq n)$
and $N$ is the set of players in each of the games in the
sequence. The game to be played in iteration $i$, $g_i$, is
randomly selected from the set $G$ independently of previous
selections made. The strategy of agent $t$ in $\bar{G}$ determines
its strategy in  $g_i$, $1\leq i \leq n$, as a function of the
strategies of $g_j$, $1 \leq j < i$, selected  by the other agents in
previous iterations. Given a tuple of strategies of $\bar{G}$,
one for each agent, the payoff of agent $t$ is taken as the sum
of its payoffs in the $n$ iterations.
A sub-game of a random game $\bar{G}$ with respect to $G$, is
a random game with respect to $G$ that starts from iteration $1 \leq j \leq n$
and consists of $n-j+1$ iterations as above.
A (rationally) competitive strategy in a random game is required to be
a (rationally) competitive strategy at each sub-game of it. 
}

\begin{theorem}

Given a random game $\bar{G}$ with  respect to
$G=\{G_1,G_2,\ldots,G_m\}$, and assume that
 $h_i(G_k) \geq 2 \cdot sh_i(G_l)$ for every $1 \leq k,l \leq m$, 
and for every agent $i$, then
a rationally competitive strategy for agent $i$ in the game
$\bar{G}$ is obtained
 by performing the competitive strategy of game $g_i$ in 
 iterations
 $1,2,\ldots,n-1$ and performing the rational competitive strategy
 of the game $g_n$ on the last iteration in the sequence. 
\end{theorem}

\section{One-way trading in multi-agent systems}

In the previous section we have discussed competitive analysis
for multi-agent systems in the framework of general repeated games
and random games. In this section we look at a particular variant 
of repeated games that extends a well known and fundamental framework
for competitive analysis -- the one-way trading (see citations 
in chapter 14 of \cite{BorElYa}).

One way to present the structure of one-way trading is as
follows. An agent $a$ seeks buying $X$ units of a good or of a
service. A supplier $A$ wishes to supply these units of good to
$a$. The agents act in an environment that determines the actual
payment for a unit of good in a non-deterministic way. For example,
the payments might be specified in dollars, but since agent $A$ is
a British company the actual payoffs it will obtain for providing 
the good  will depend on
the exchange ratio of the dollar and the 
British pound. Formally, the environment
announces at each point in time, $1,2,\ldots,t$,  the payoff that
will be obtained by agent $A$ for supplying a  unit of good. The
announcements are selected in an unpredicted non-deterministic
manner from the interval $[m,M]$, where $M > m
> 0$. For example,  when  $K$ is announced at point $i$,
agent $A$ can supply the $X$ units of good and obtain 
a payoff of $X \cdot K$. Our assumption is that agent $A$ will
obtain a zero payoff by not providing the units of good.
 The
decision problem that agent $A$ faces is as follows: at each point
he needs to decide whether he would like to supply the
units of good in the current rate. We assume that when agent $A$
is willing to provide the service then he will provide and be paid
for the whole quantity of goods requested  by agent $a$ (this property
is termed one-way search).

The competitive analysis approach tells agent $A$ in the above
scenario to minimize his regret value. As it turns out, the
competitive strategy in this case will tell the agent to accept
the offer (i.e supply the units of good) when the payoff
reaches ${{M-m} \over 2}$ in stage $j\leq t-1$, and to accept 
the offer on stage $t$ otherwise. 

One-way trading is a typical setting for the use of competitive
analysis. We now extend it to the case of several agents, where
more than one agent may wish to supply the units of good
requested by $a$. We will first develop the multi-agent 
framework without
considering the rationality assumption, and then will extend it to
the case of rational competitive analysis.

\subsection{Multi-agent one-way trading}

For ease of exposition we consider the case of trading two
agents (i.e. two suppliers who can provide the units of good
requested by $a$): $A_1$ and $A_2$. The payment offers for the
two agents are taken to be independent. For example, agent 1 may be a 
British company and agent 2 may be a Japanese company, and
therefore the actual payment offers for them (from their
perspective) will reflect the exchange ratio between the dollar
and the British pound, and the exchange ratio between the dollar and
the Japanese Yen, respectively. Formally, we have:

\begin{definition}
Let $M_1,M_2,m_1,m_2,t,X=2K$ be positive 
integers, where $M_1 > m_1$ and $M_2
> m_2$, $t \geq 3$, and $X$ is even. A multi-agent one way trading $T=\langle
N=\{1,2\},X,t,m_1,M_1,m_2,M_2  \rangle$ is a random game with the
following players, strategies,  and payoffs:

\begin{enumerate}
\item
The players are 1 and 2.
\item There are $t$ iterations. Each iteration $i$ is associated with
a pair of numbers $(a_1,a_2)$ where $m_1 \leq a_1 \leq M_1$ and
$m_2 \leq a_2 \leq M_2$. At each iteration each agent can "take"
or "pass". However, if an agent takes in iteration $j$ then both
agents  can only "pass" in all iterations $j \geq i$.
\item The payoff of each agent in iteration $i$ is 0 if it passes;
if an agent performs "take" in iteration $i$ then its payoff will
be $a_i X$ if the other agent passes and $a_i K$ if the other
agent takes.
\end{enumerate}
\end{definition}

Intuitively, "take" means a decision of accepting the offer,
while "pass" means rejecting it (at the given point). If both
agents agree to "take" then each one of them will supply half of
the units (and the payoff will be splitted among the agents).
 We
now show what is the structure of the competitive strategy in a
multi-agent one-way trading setting.

\begin{theorem}
Given a multi-agent one way trading $T=\langle
N=\{1,2\},X,t,m_1,M_1,m_2,M_2  \rangle$, a competitive strategy for agent $i$
is as follows:

\begin{enumerate}
\item For iterations $1 \leq j \leq t-1$, take iff $a_i \geq
{{2M_i+m_i} \over 4}$
\item If you arrive at iteration $t$ then take.
\end{enumerate}
\end{theorem}

\noindent{Basic idea behind proof:}
Consider iteration $j$, $1 \leq j \leq t-1$, and consider the announcement
$a_i=Y$. Then, by taking in round $j$, agent $i$ might suffer a regret of
$2M_iK-2YK$ (notice that there is a regret when an agent takes only 
if the other does not take at that iteration). 
By not taking in stage $j$  agent $i$ might suffer a regret
of $2YK-m_iK$ (which is in fact $max(KY,2YK-m_iK)$). 
 In order to minimize the regret agent $i$ will therefore have
to take whenever $Y$ satisfies that $2M_iK-2YK=2YK-m_iK$, i.e. when
$a_i={{2M_i+m_i}\over 4}$. The fact that the regret is minimized in iteration
$t$ by taking rather than passing is immediate.

\subsection{Rational competitive analysis for multi-agent one way
trading}

We now show the result
of applying rational competitive analysis to the context of
multi-agent one way trading:

\begin{theorem}
Given a multi-agent one way trading $T=\langle
N=\{1,2\},X,t,m_1,M_1,m_2,M_2  \rangle$, a rational
competitive strategy for agent $i$
is as follows:

\begin{enumerate}
\item For iterations $1 \leq j \leq t-1$, if the other agent, $k$, is
announced that $a_k=M_k$, then take.

\item For iterations $1 \leq j \leq t-2$, if (1) does not hold then
take iff $a_i \geq
{{2M_i+m_i} \over 4}$

\item If (1) does not hold, then in iteration $t-1$
take iff $a_i \geq
 {{M_i+m_i} \over 4}$.
\item If you arrive at iteration $t$ then take.
\end{enumerate}
\end{theorem}

\noindent{Basic idea behind proof:}
Notice that if the other agent, $k$, is announced that $a_k=M_k$ then taking
dominates any other strategy of it. In no other cases we can say that taking
or passing in iterations $1,2,\ldots,t-1$ is dominated. Also, in stage $t$
passing is dominated by taking.
As a result we will get that agent $i$ will minimize its regret by taking
when $a_k=M_k$ or when it arrived in the last iteration.
Assume that $a_i=Y$ in iteration $t-1$, then
 the maximal regret we get by taking  is $M_iK-2KY$, and by passing
the maximal regret in this case  is $2KY-m_iK$ (which is in fact 
$max(KY,2KY-m_iK)$).  This implies that 
the regret is minimized 
when $M_iK-2YK=2KY-m_iK$ ), i.e. 
when $Y = {{M_i+m_i} \over 4}$.
The other case that refers to iterations $1,2,\ldots,t-2$ will be treated as
in the case of (standard) competitive analysis.


\section{Conclusion}

Competitive analysis
is a major tool in computer science, which has been used in a variety
of contexts.
In this paper we have introduced rational competitive
analysis. Rational competitive analysis generalizes competitive
analysis to the context of multi-agent systems. 
 Moreover, we have shown its use in the context of bidding games 
and one-way trading, two problems of considerable importance, 
as well as in the context of general repeated games. 
 Our approach adopts the non-Bayesian normative
approach adopted in previous work, but modifies it to incorporate
minimal rationality requirements. Such requirements are essential in
multi-agent domains.
 Many of the previous studies in the context of  competitive analysis 
 can be naturally extended to multi-agent 
domains, and then rational competitive analysis can  serve as a fundamental 
tool for the study of these extensions. 
  We see the study of such extensions as a most attractive 
research topic, and hope 
that others will join us in addressing this challenge.

\end{document}